\documentclass[10pt,twocolumn,letterpaper]{article}
\newcommand{\final}{0}
\usepackage{cvpr}
\usepackage{xcolor}
\usepackage{times}
\usepackage{epsfig}
\usepackage{graphicx}
\usepackage{amsmath}
\usepackage{amssymb}
\usepackage{multirow}
\usepackage{booktabs}
\usepackage{subfigure}
\usepackage{microtype}
\usepackage{ifthen}
\definecolor{WeimingColor}{rgb}{0,0,0.8}
\definecolor{XingjiaColor}{rgb}{0.0,0.1,0.9}
\definecolor{FanColor}{rgb}{0.8,0,0.8}
\definecolor{SaulColor}{rgb}{0.8,0.1,0}
\definecolor{CondiColor}{rgb}{0.0,0.8,0.4}
\definecolor{ChongyangColor}{rgb}{0.8,0,0.8}

\newcommand{\weiming}[1]{{\color{WeimingColor} [Weiming: #1]}}
\newcommand{\xingjia}[1]{{\color{XingjiaColor}[Xingjia: #1]}}
\newcommand{\saul}[1]{{\color{SaulColor}[Saul: #1]}}
\newcommand{\condi}[1]{{\color{CondiColor}[Condi: #1]}}
\newcommand{\fan}[1]{{\color{FanColor}[Fan: #1]}}
\newcommand{\chongyang}[1]{{\color{ChongyangColor}[Chongyang: #1]}}

\newcommand{\warning}[1]{{\it\color{red} #1}}
\newcommand{\toremove}[1]{{\it\color{red} (To remove) #1}}
\newcommand{\note}[1]{{\it\color{blue} #1}}
\newcommand{\nothing}[1]{}

\ifthenelse{\equal{\final}{1}}
{
\renewcommand{\weiming}[1]{}
\renewcommand{\fan}[1]{}
\renewcommand{\xingjia}[1]{}
\renewcommand{\saul}[1]{}
\renewcommand{\condi}[1]{}
\renewcommand{\chongyang}[1]{}

\renewcommand{\warning}[1]{}
\renewcommand{\toremove}[1]{}
\renewcommand{\note}[1]{}
\renewcommand{\nothing}[1]{}
}{}

\usepackage[pagebackref=true,breaklinks=true,letterpaper=true,colorlinks,bookmarks=false]{hyperref}

\cvprfinalcopy 

\def\cvprPaperID{8906} 

\ifcvprfinal\fi
\begin{document}

\title{Dynamic Refinement Network for Oriented and\\ Densely Packed Object Detection}

\author{Xingjia Pan$^{1,2}$\quad Yuqiang Ren$^{3}$\quad Kekai Sheng$^{3}$\quad Weiming Dong$^{1,2,4}$\thanks{Corresponding author}  \\
Haolei Yuan$^{3}$\quad Xiaowei Guo$^{3}$\quad Chongyang Ma$^{5}$\quad Changsheng Xu$^{1,2,4}$\\
$^1$NLPR, Institute of Automation, CAS\quad $^2$School of Artificial Intelligence, UCAS \\$^3$Youtu Lab, Tencent\quad $^4$CASIA-LLVision Joint Lab\quad $^5$Y-Tech, Kuaishou Technology\\
{\tt\small \{panxingjia2015, weiming.dong, changsheng.xu\}@ia.ac.cn}, {\tt\small chongyangma@kuaishou.com} \\
{\tt\small \{condiren, saulsheng, harryyuan, scorpioguo\}@tencent.com}
}

\maketitle

\begin{abstract}
Object detection has achieved remarkable progress in the past decade. However, the detection of oriented and densely packed objects remains challenging because of following inherent reasons: (1) receptive fields of neurons are all axis-aligned and of the same shape, whereas objects are usually of diverse shapes and align along various directions; (2) detection models are typically trained with generic knowledge and may not generalize well to handle specific objects at test time; (3) the limited dataset hinders the development on this task.
To resolve the first two issues, we present a dynamic refinement network which consists of two novel components, i.e., a feature selection module (FSM) and a dynamic refinement head (DRH).
Our FSM enables neurons to adjust receptive fields in accordance with the shapes and orientations of target objects, whereas the DRH empowers our model to refine the prediction dynamically in an object-aware manner.
To address the limited availability of related benchmarks, we collect an extensive and fully annotated dataset, namely, SKU110K-R, which is relabeled with oriented bounding boxes based on SKU110K.
We perform quantitative evaluations on several publicly available benchmarks including DOTA, HRSC2016, SKU110K, and our own SKU110K-R dataset.
Experimental results show that our method achieves consistent and substantial gains compared with baseline approaches.
The code and dataset are available at \url{https://github.com/Anymake/DRN_CVPR2020}.
\end{abstract}

\section{Introduction}
\begin{figure}
\centering
\subfigure[Classification]{
\includegraphics[width=0.45\linewidth]{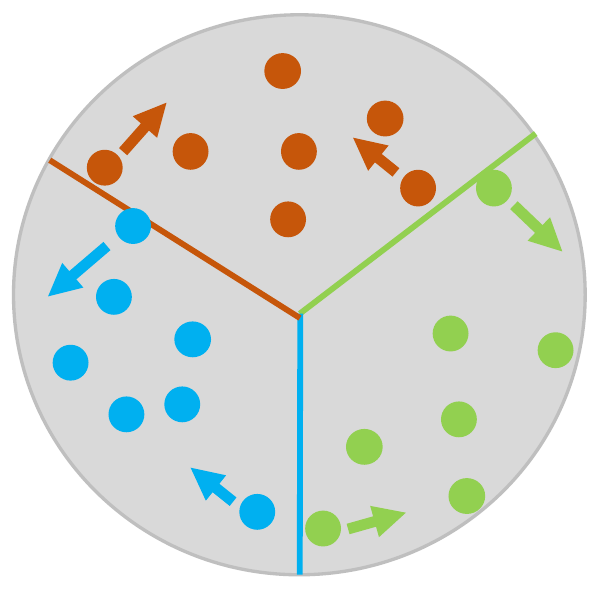}
\label{fig:crc}}
\subfigure[Regression]{
\includegraphics[width=0.49\linewidth]{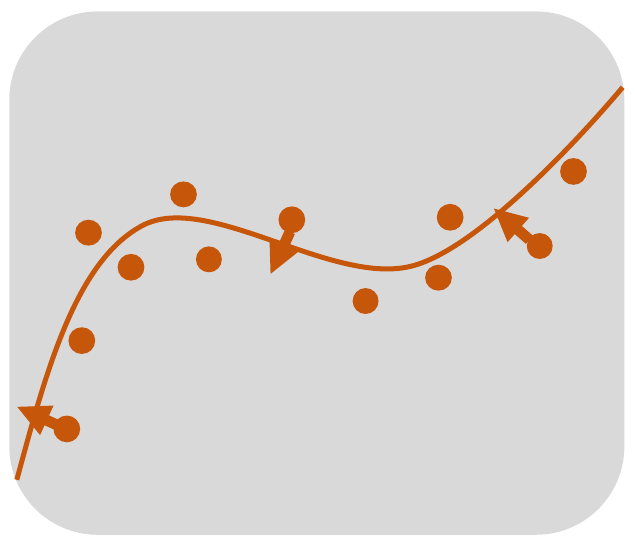}
\label{fig:crr}}
\caption{Illustrations of dynamic refinement on classification (a) and regression (b).
Each solid dot represents a sample. 
With the general knowledge learned in training procedure, classifiers and regressors make predictions while suffering from lack of flexibility. Model should changes over samples. The arrows show the promising refinements for improved performance.}
\label{fig:cr}
\end{figure}
Object detection has achieved remarkable progress on a few benchmarks (e.g., VOC~\cite{everingham2010pascal} and COCO~\cite{Lin:2014:COCO}) with the help of deep learning. Numerous well-designed methods~\cite{ren2015faster,zhou2019objects,zhu2019feature,redmon2018yolov3,Borji:2019:SOD} have demonstrated promising results. However, majority of these detectors encounter problems when objects, such as those in aerial images, are in arbitrary orientations and present dense distribution.
Moreover, almost all detectors optimize model parameters on the training set and keep them fixed afterward.
This static paradigm, which uses general knowledge, may not be flexible enough to detect specific samples during test time.

Most of the recent progress on oriented object detection is based on R-CNN series frameworks~\cite{girshick2014rich,girshick2015fast,ren2015faster}. These methods first generate numerous horizontal bounding boxes as region of interests (RoIs) and then predict classification and location on the basis of regional features.
Unfortunately, horizontal RoIs typically suffer from severe misalignment between the bounding boxes and \emph{oriented} objects~\cite{xia2018dota,liu2016ship}.
For example, objects in aerial images are usually with arbitrary orientations and densely packed, leading to artifacts wherein several instances are often crowded and contained by a single horizontal RoI~\cite{ding2019learning}. Consequently, extracting accurate visual features becomes difficult. Other methods~\cite{xia2018dota, liu2015fast,liu2016ship,liu2017rotated} leverage oriented bounding boxes as anchors to handle rotated objects.
However, these methods suffer from high computational complexity because they acquire numerous well-designed anchors with different angles, scales, and aspect ratios.
Recently, RoI Trans~\cite{ding2019learning} has transformed horizontal RoIs into oriented ones by rotating RoI learners and extracting rotation-invariant region features using a rotated position-sensitive RoI alignment module.
However, such approach still needs well-designed anchors and is not flexible enough.

\begin{figure*}
\centering
\includegraphics[width=\linewidth]{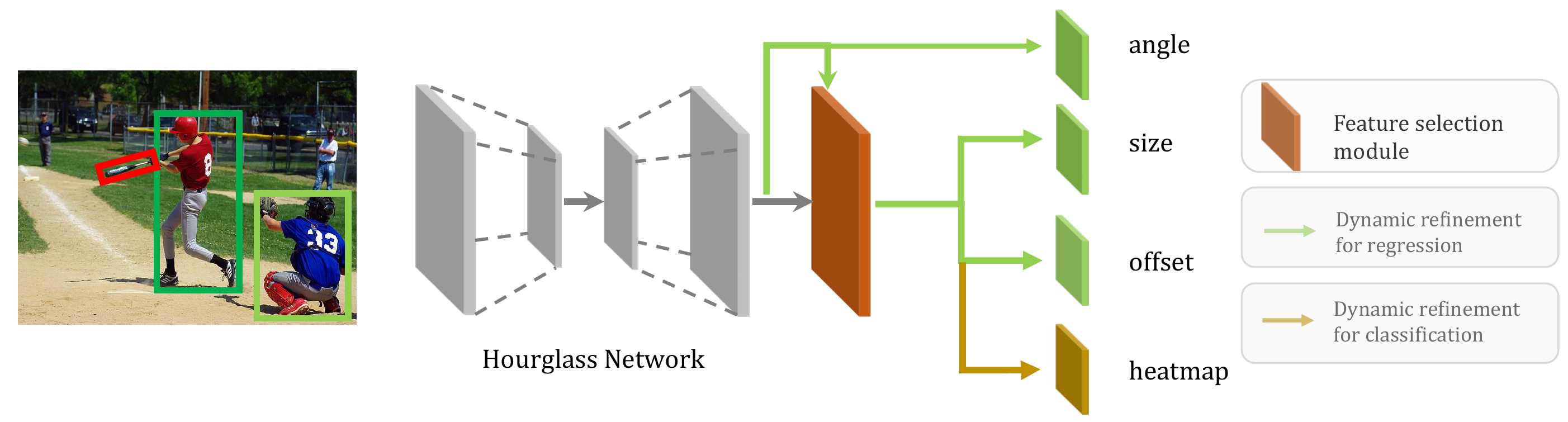}
\vspace{-15pt}
\caption{Overall framework of our Dynamic Refinement Network. The backbone network is followed by two modules, \textit{i.e.}, feature selection module (FSM) and dynamic refinement heads (DRHs).
FSM selects the most suitable features by adaptively adjusting receptive fields.
The DRHs dynamically refine the predictions in an object-aware manner.}
\label{fig:overview}
\end{figure*}

Model training is a procedure from special to general, whereas inference is from general to special.
However, almost all methods follow the stationary paradigm and cannot make flexible inference based on samples. Dynamic filters are a simple yet effective approach to enable the model to change over different samples.
Existing methods~\cite{dai2017deformable,wang2019carafe} resort to feature reassembly via dynamic filters and achieve promising results.
However, detectors have two different tasks, namely, classification and regression.
Fig.~\ref{fig:cr} shows some illustrative examples.
For a classification task, the key is to refine the feature embedding for improved discriminability.
However, for a regression problem, refining the predicted value directly is desirable.
We propose two versions of dynamic refinement heads (DRHs) tailored for the above two aspects.

In this work, we adopt CenterNet~\cite{zhou2019objects}, with an additional angle prediction head as our baseline and present dynamic refinement network (DRN).
Our DRN consists of two novel parts: feature selection module (FSM) and dynamic refinement head (DRH).
FSM empowers neurons with the ability to adjust receptive fields in accordance with the object shapes and orientations, thus passing accurate and denoised features to detectors.
DRH enables our model to make flexible inferences in an object-aware manner.
Specifically, we propose two DRHs for classification (DRH-C) and regression (DRH-R) tasks.
In addition, we carefully relabel oriented bounding boxes for SKU110K~\cite{goldman2019precise} and called them SKU110K-R; in this manner, oriented object detection is facilitated.
To evaluate the proposed method, we conduct extensive experiments on the DOTA, HRSC2016, and SKU110K datasets.

In summary, our contributions include:
\begin{itemize}
\setlength{\itemsep}{0pt}
\setlength{\parskip}{0pt}
\setlength{\parsep}{0pt}
    \item We propose a novel FSM to adaptively adjust the receptive fields of neurons based on object shapes and orientations. The proposed FSM effectively alleviates the misalignment between receptive fields and objects.
    \item We present two DRHs, namely, DRH-C and DRH-R, for classification and regression tasks, respectively. These DRHs can model the uniqueness and particularity of each sample and refine the prediction in an object-wise manner.
    \item We collect a carefully relabeled dataset, namely, \emph{SKU110K-R}, which contains accurate annotations of oriented bounding boxes, to facilitate the research on oriented and densely packed object detection.
    \item Our method shows consistent and substantial gains across DOTA, HRSC2016, SKU110K, and SKU110K-R on oriented and densely packed object detection.
\end{itemize}

\section{Related Work}
\label{sec:rw}

\nothing{Object detection methods can be roughly divided into two kinds today: top-down and bottom-up. The majority of methods follow top-down paradigm~\cite{ren2015faster,Liu:2016:SSD,Redmon:2016:YOLO}. They usually predict results based on pre-defined anchors or proposals.
R-CNN series work~\cite{girshick2014rich,girshick2015fast,ren2015faster} are the cradle of two-stage methods, which usually first aim to attain refined proposals through RPN~\cite{ren2015faster} and further outputs the detection results in the form of bounding boxes by using a small detector. One-stage methods, like SSD variants~\cite{Liu:2016:SSD,Shen:2017:DSOD}, YOLO~\cite{Redmon:2016:YOLO,Redmon:2017:Yolo9000,redmon2018yolov3} variants and RetinaNet~\cite{lin2017focal}, predict the detection results directly from anchors after extracting features from the image. Bottom-up approaches~\cite{law2018cornernet,zhou2019objects,duan2019centernet} try to acquire detection results by grouping sub-object entities like keypoints or parts.  Most of these methods focus on proceeding horizontal object detection.}
Most object detection methods~\cite{ren2015faster,Liu:2016:SSD,Redmon:2016:YOLO,Shen:2017:DSOD,redmon2018yolov3,law2018cornernet,zhou2019objects,Song:2019:TSR} focus on axis-aligned or upright objects and may encounter problems when the targets are of arbitrary orientations or present dense distribution~\cite{goldman2019precise}.
For oriented object detection, some methods \cite{girshick2014rich,hsieh2017drone,lin2014microsoft,liu2016ship, liu2017rotated} adopt the R-CNN \cite{ren2015faster} framework and use numerous anchors with different angles, scales, and aspect ratios, at the expense of considerably increasing computation complexity.
The SRBBS~\cite{liu2016ship} uses rotated region of interest (RoI) warping to extract features of rotated RoIs; however, it is difficult to embed in a neural network because rotated proposal generation consumes additional time.
Ding \etal~\cite{ding2019learning}  proposed an RoI transformer to transform axis-aligned RoIs into rotated ones to address the misalignment between RoIs and oriented objects. SCRDet~\cite{yang2019scrdet} added an IOU constant factor to the L1 loss term to address the boundary issue for oriented bounding boxes.
In contrast to the aforementioned methods, we propose FSM to adjust receptive fields of neurons adaptively and reassemble appropriate features for various objects with different angles, shapes, and scales.

FPN~\cite{lin2017feature} proposes a feature pyramid network to perform object detection at multiple scales. They select features of the proposals in accordance with area sizes.
FSAF~\cite{zhu2019feature} learns an anchor-free module to select the most suitable feature level dynamically.
Li \etal~\cite{li2019dynamic} presented a dynamic feature selection module to select pixels on basis of the position and size of new anchors.
These methods aim to select additional suitable features at the object level.
To become more fine-grained, SKN~\cite{li2019selective} learned to select features with different receptive fields at each position using different kernels.
SENet~\cite{hu2018squeeze} explicitly recalibrates channel-wise feature responses adaptively, whereas CBAM~\cite{woo2018cbam} adopts one more spatial attention module to model inter spatial relationships.
Our FSM learns to extract shape- and rotation-invariant features in a pixel-wise manner.

Spatial transformer network~\cite{jaderberg2015spatial} are the first to learn spatial transformation and affine transformation in deep learning frameworks to warp feature maps.\nothing{, which is known difficult to learning}
Active convolution~\cite{jeon2017active} augments the sampling locations in the convolutional layers with offsets. It shares the offsets all over the different spatial locations and the model parameters are static after training.
Deformable convolutional network (DCN)~\cite{dai2017deformable} models the dense spatial transformation in the images and the offsets are dynamic model outputs.
Our rotated convolution layer in FSM learns the rotation transformation in a dense fashion\nothing{ and it is also lightweight and dynamic}.
RoI Trans~\cite{ding2019learning} learns five offsets to transform the axis-aligned ROIs into rotated ones in a manner similar to that of position-sensitive ROI Align~\cite{ren2015faster}.
ORN~\cite{zhou2017oriented} proposes active rotating filters which actively rotate during convolution. The rotation angle is a hyper-parameter which is rigid and all the locations share the same rotation angle.
On the contrary, our rotation transformation is learnable and can predict angles at each position.

Neural networks are conditioned on the input features and change over samples by introducing dynamic filters.
Dynamic filters~\cite{jia2016dynamic} learns filter weights in the training phase and thus can extract example-wise features at the inference time.
Similarly, CARAFE~\cite{wang2019carafe} proposes a kernel prediction module which is responsible for generating the reassembly kernels in a content-aware manner.
Although DCN~\cite{dai2017deformable} and RoI Trans~\cite{ding2019learning} model the offset prediction in a dynamic manner, they do not change the kernel weight.
In contrast to~\cite{dai2017deformable,wang2019carafe}, our DRHs aim to refine the detection results in a content-aware manner by introducing dynamic filters instead of feature reassembly.

\section{Our Method and Dataset}
\label{sec:drn}
The overall framework of our approach is shown in Fig.~\ref{fig:overview}.
We first introduce our network architecture in Sec.~\ref{sec:na}. The misalignment between various objects and simplex receptive fields in each network layer is ubiquitous; hence, we propose an FSM to reassemble the most suitable feature automatically, as described in Sec.~\ref{sec:fsm}.
To empower a model with the ability to refine predictions dynamically in accordance with different examples, we propose the use of DRHs to achieve object-aware predictions in Sec.~\ref{sec:drh}.

\subsection{Network Architecture}
\label{sec:na}
We use CenterNet~\cite{zhou2019objects} as our baseline, which models an object as a single point (i.e., the center point of the bounding box) and regresses the object size and offset.
To predict oriented bounding boxes, we add a branch to regress the orientations of the bounding boxes, as illustrated in Fig.~\ref{fig:overview}.
Let $(c_x, c_y, h, w, \theta, \delta_x, \delta_y)$ be one output septet from the model.
Then, we construct the oriented bounding box by:
\begin{equation}
  \begin{aligned}
      P_{lt} &= M_r [-w/2, -h/2]^T + [c_x+\delta_x,  c_y+\delta_y]^T, \\
      P_{rt} &= M_r [+w/2, -h/2]^T + [c_x+\delta_x,  c_y+\delta_y]^T, \\
      P_{lb} &= M_r [-w/2, +h/2]^T + [c_x+\delta_x,  c_y+\delta_y]^T, \\
      P_{rb} &= M_r [+w/2, +h/2]^T + [c_x+\delta_x,  c_y+\delta_y]^T,
    \label{Equ:consbb}
\end{aligned}
\end{equation}
where $(c_x, c_y)$ and $(\delta_x, \delta_y)$ are the center point and the offset prediction; $(w,h)$ is the size prediction; $M_r$ is the rotation matrix; and $P_{lt}$, $P_{rt}$, $P_{lb}$ and $P_{rb}$ are the four corner points of the oriented bounding box.
Following CenterNet for regression tasks, we use L1 loss for the regression of rotation angles:
\begin{equation}
    L_{ang} = \frac{1}{N}\sum_{k=1}^N|\theta - \hat{\theta}|,
    \label{Equ:loss_angle}
\end{equation}
where $\theta$ and $\hat{\theta}$ are the target and predicted rotation angles, respectively; and $N$ is the number of positive samples. Thus, the overall training objective of our model is
\begin{equation}
    L_{det} = L_k + \lambda_{size}L_{size} + \lambda_{off}L_{off} + \lambda_{ang}L_{ang},
    \label{Equ:total_loss}
\end{equation}
where $L_k$, $L_{size}$ and $L_{off}$ are the losses of center point recognition, scale regression, and offset regression, which are the same as CenterNet; and $\lambda_{size}$, $\lambda_{off}$ and $\lambda_{ang}$ are constant factors, which are all set to $0.1$ in our experiments.

\subsection{Feature Selection Module}
\label{sec:fsm}
\begin{figure}
\centering
\includegraphics[width=\linewidth]{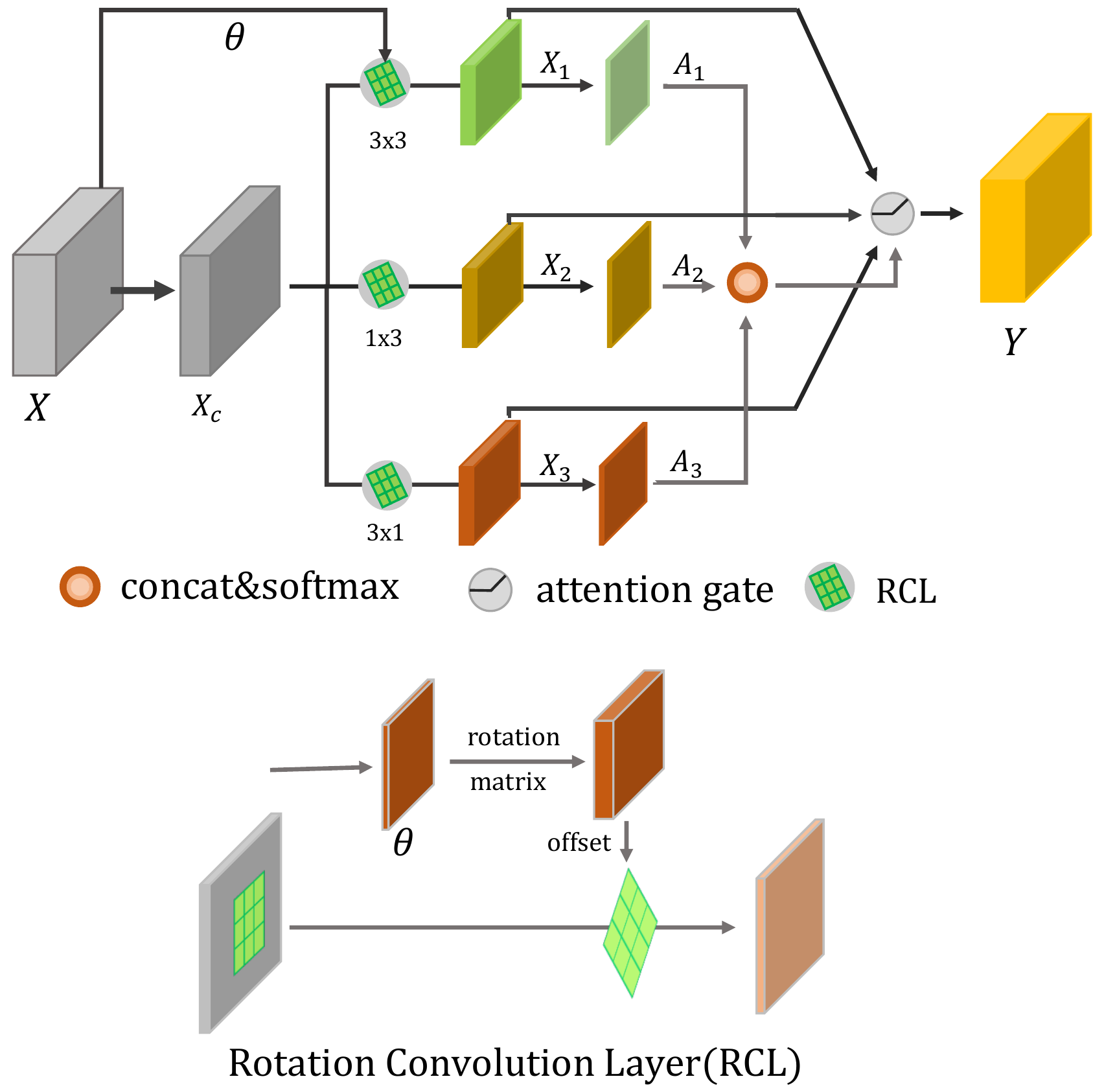}
\caption{\textbf{Top}: Feature Selection Module. \textbf{Bottom}: Rotation Convolution Layer. The illustration shows a three-split example.
Each split extracts different information by using Rotation Convolution Layer with $3\times3$, $1\times3$, and $3\times1$ kernels.
We adopt the attention mechanism to aggregate the information.
}
\label{fig:fsm}
\end{figure}

To alleviate the mismatches between various objects and axis-aligned receptive fields of neurons, we propose an Feature Selection Module (FSM) to aggregate the information extracted using different kernel sizes, shapes (aspect ratios), and orientations adaptively (see Fig. \ref{fig:fsm}).

\paragraph{Multiple features.}
Given a feature map $X$ $\in$ $\mathbb{R}^{H \times W \times C}$, we first compress the feature with a $1 \times 1$ convolution layer, followed by Batch Normalization\cite{ioffe2015batch} and ReLU\cite{nair2010rectified} function in sequence for improved information aggregation.
Next, we extract multiple features by using Rotation Convolution Layers (RCLs) with different kernels from $X_c$ $\in$ $\mathbb{R}^{H \times W \times C^{'}}$. Fig.~\ref{fig:fsm} shows a three-split example with $3\times3$, $1\times3$, and $3\times1$ kernels. Each split is responsible for different receptive fields, and we call it $X_i$ $\in$ $\mathbb{R}^{H \times W \times C^{'}}$, where $i$ $\in$ $\{1,2,3\}$. The RCL draws inspiration from DCN~\cite{dai2017deformable}, and the implementation details are shown in Fig.~\ref{fig:fsm}.
Akin to DCN, we use $\mathcal{R}$ to represent the regular grid receptive field and dilation. For a kernel of size $3 \times 3$, we have
\begin{equation}
    \mathcal{R}=\{(-1, -1), (-1, 0), ... ,(0, 1), (1, 1)\}.
\end{equation}
Given the pre-defined offset $p_i$ $\in$ $R$ for the $i$-$th$ location and learned angle $\theta$, the learned offset is
\begin{equation}
  \begin{aligned}
      \delta p_i &= M_r(\theta) \cdot p_i - p_i,
    \label{Equ:off}
\end{aligned}
\end{equation}
where $M_r(\theta)$ is the rotation matrix defined in Eqn.~(\ref{Equ:consbb}).
For each location $p_0$ in the output feature map $X_i$, we have
\begin{equation}
    X_i(p_0) = \sum_{p_n\in \mathcal{R}}w(p_n)\cdot X_c(p_0+p_n+\delta p_n),
\end{equation}
where $p_n$ denotes the locations in $\mathcal{R}$, and $w$ is the kernel weight.

\paragraph{Feature selection.}
To enforce neurons with adaptive receptive fields, we adopt an attention mechanism to fuse the feature in a position-wise manner. $X_i$ is first to feed into an attention block (composed of a convolution with kernel $1\times1$, Batch Normalization and ReLU in sequence) to obtain the attention map $A_i$ $\in$ $R^{H \times W \times 1}$ ($i$ $\in$ ${1,2,3}$).
Then, we concatenate $A_{i}$ in the channel direction, followed with a SoftMax operation to obtain the normalized selection weight $A^{'}_{i}$ as:
\begin{equation}
    A^{'}_{i} = SoftMax([A_{1}, A_{2}, A_{3}]).
    \label{Equ:watt}
\end{equation}
A soft attention fuses features from multiple branches:
\begin{equation}
    Y = \sum_{i}A^{'}_{i}\cdot X_{i},
    \label{Equ:fuse}
\end{equation}
where $Y$ $\in$ $\mathbb{R}^{H \times W \times C}$ is the output feature.  We omit the channel expansion layer before $Y$ for similarity.
Here, we show a three-branch case, and one can easily extend to more branches with different kernel sizes and shapes.

\subsection{Dynamic Refinement Head}
\label{sec:drh}
In standard machine learning frameworks, people usually learn a model through a large annotated training set.
At the inference time, the test example is fed to the model with parameters fixed to obtain the prediction results.
A problem occurs when the well-trained model can only respond on the basis of the general knowledge learned from the training set while ignoring the uniqueness of each example.

To enable the model to respond on the basis of each sample, we propose the use of DRHs to model the particularity of each input object. Specifically, two different modules, \emph{i.e.,} DRH-C and DRH-R, can be used for classification and regression, respectively.

\begin{figure}
\centering
\includegraphics[width=\linewidth]{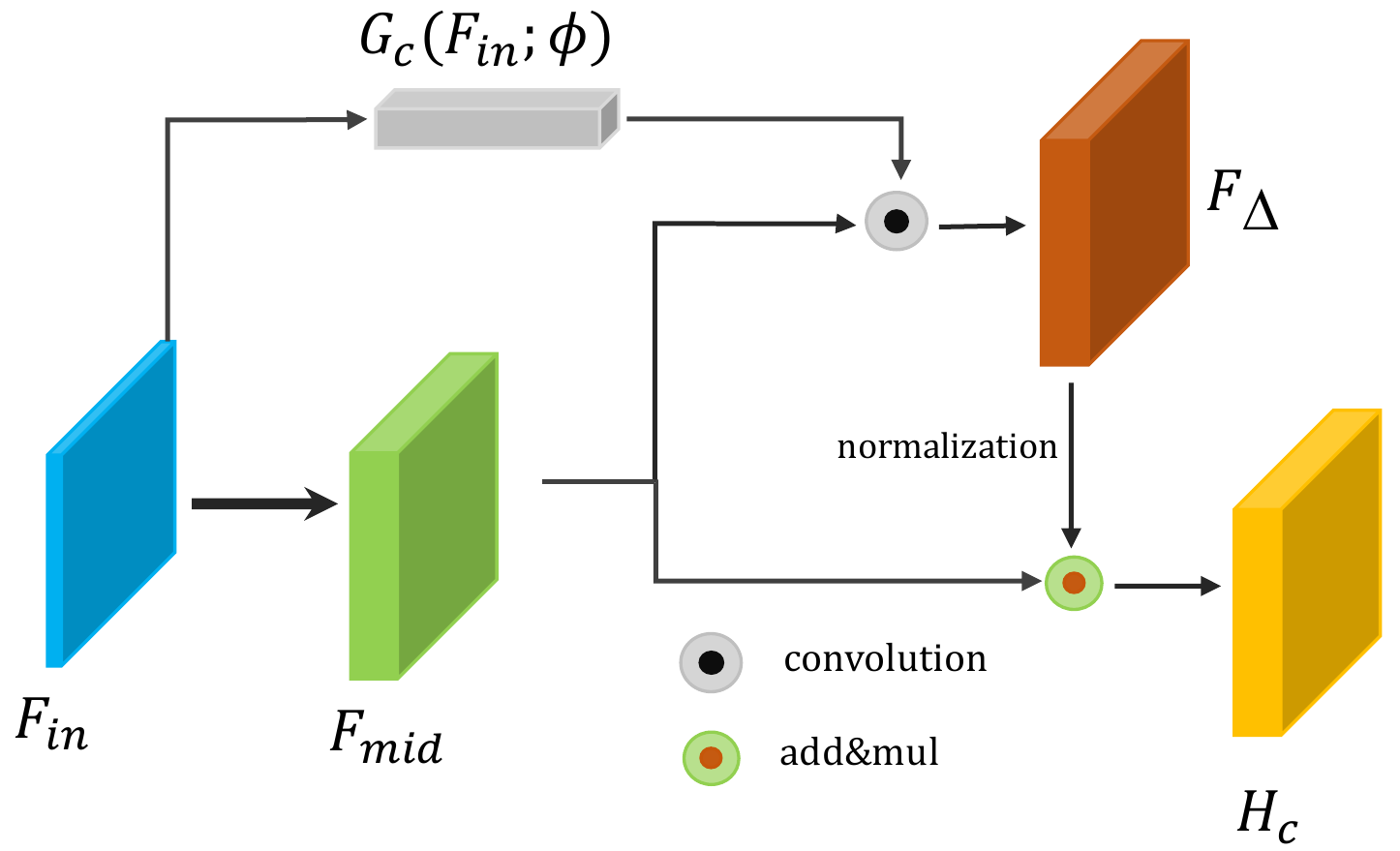}
\caption{Dynamic Refinement Head for classification (DRH-C).}
\label{fig:dfmc}
\end{figure}

We illustrate our motivation with an example for a three-class classification problem, as shown by the left image in Fig.~\ref{fig:crc}. The gray circular area represents the feature space and solid dots are examples that belong to three classes. Some samples are located far from the discrimination boundary, indicating that these samples possess good semantic discriminability. By contrast, the samples with a small margin to the boundary are unfortunately not much compatible with the model. To enhance the flexibility of the model, we resort to an object-aware classification/regression module.

\paragraph{Dynamic refinement for classification.}
The architecture of DRH-C is shown in Fig.~\ref{fig:dfmc}.
Given an input feature map $F_{in}$ $\in$ $R^{H \times W \times C}$, we first obtain an object-aware filter:
\begin{equation}
     K_c= G_c(F_{in}; \phi),
    \label{Equ:mc_kg}
\end{equation}
where $G_c$ represents the dynamic filter generator, and $\phi$ is the parameter set of $G_c$. $K_c$ are the learned example-wise kernel weights.
Then, we obtain the feature refinement $F_{\vartriangle}$ via a convolution operation:
\begin{equation}
     F_\vartriangle= F_{mid} * K_c,
    \label{Equ:mc_conv}
\end{equation}
where $F_{mid}$ is the base feature by processing $F_{in}$ through a Conv-BN-ReLU block with kernel $3 \times 3$, and $*$ represents the convolution operator. Finally, we obtain the classification prediction $H_c$:
\begin{equation}
     H_c= C\big((1+ \varepsilon \cdot F_\vartriangle /\| F_\vartriangle \|) \cdot F_{mid}; \Phi\big),
    \label{Equ:mc_cls}
\end{equation}
where $C(\cdot, \Phi)$ represents the classifier with parameter $\Phi$, and $\|\cdot \|$ is a modulus operation. We normalize $F_\vartriangle$ in the channel direction for each location. The normalized $F_\vartriangle$ indicates the modification direction for base feature $F_{mid}$. We adaptively refine the basic feature according to its length. $\varepsilon$ is a constant factor to control the scope of refinement.

\begin{figure}
\centering
\includegraphics[width=\linewidth]{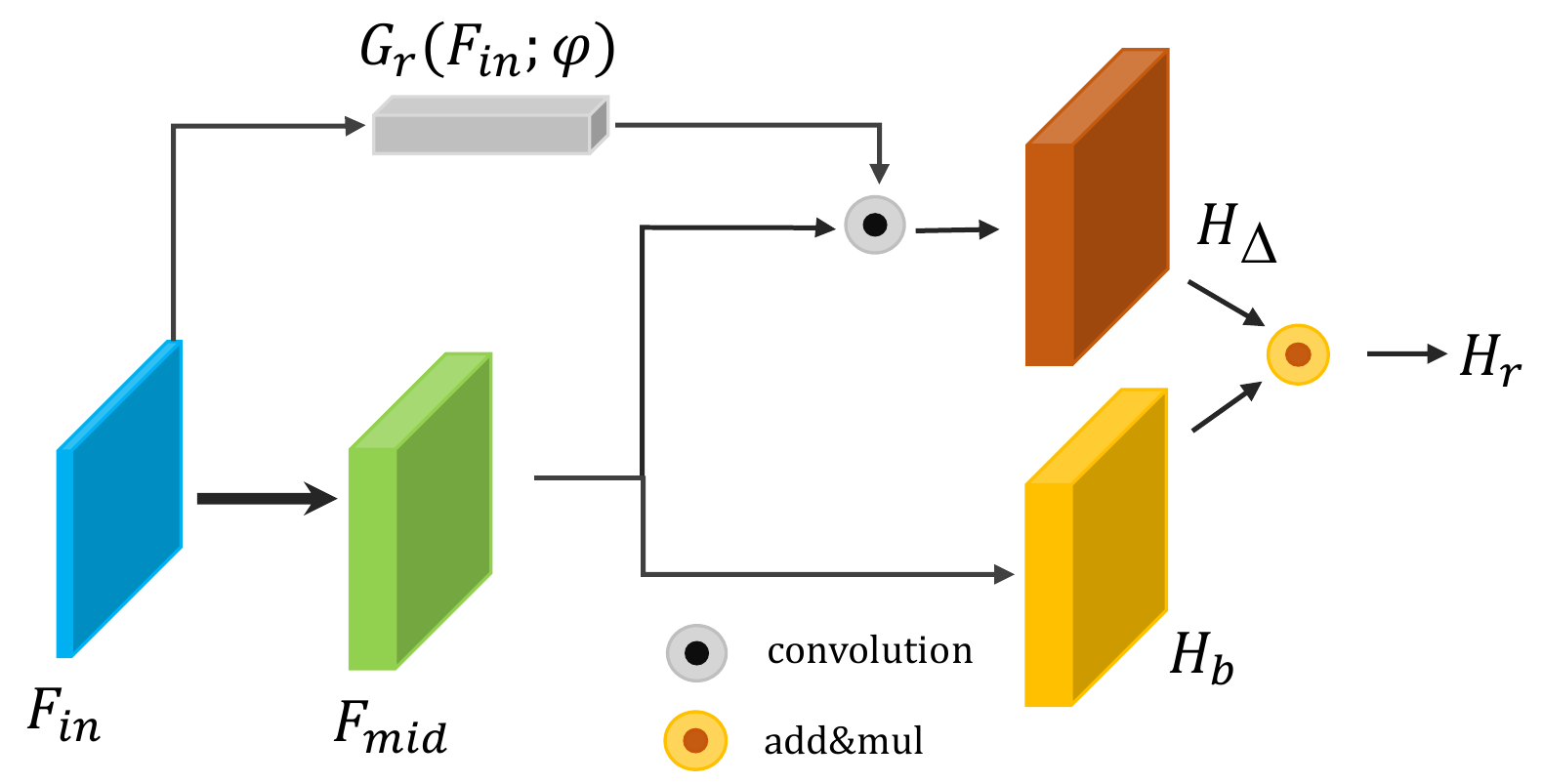}
\caption{Dynamic Refinement Head for regression (DRH-R).}
\label{fig:dfmr}
\end{figure}

\paragraph{Dynamic refinement for regression.}
We also show a simple example for regression tasks in Fig.~\ref{fig:crr}. The orange solid dots represent the target values of examples, and the orange curve represents the learned regression model. For regression tasks, researchers usually minimize the average L1 or L2 distances; thus, the learned model cannot fit the target value accurately. To predict exact values without increasing the risk of overfitting, we design an object-aware regression head similar to the classifier shown in Fig.~\ref{fig:dfmr}.
Given the feature map $F_{in}$ $\in$ $R^{H \times W \times C}$, we first calculate the dynamic filter weight $K_r$ via $G_r(\cdot; \varphi)$ and then predict the refinement factor $H_\vartriangle$ similar to Eqn.~(\ref{Equ:mc_conv}) to obtain the final object-aware regression result $H_r$:
\begin{equation}
    \begin{aligned}
     H_b &= R(F_{mid};\Psi), \\
     H_r &= \big(1+ \epsilon \cdot tanh(H_\vartriangle)\big) \cdot H_b,
     \end{aligned}
    \label{Equ:mc_reg}
\end{equation}
where $R(\cdot; \Psi)$ is the regressor with parameters $\Psi$. $H_b$ is the base prediction value according to the general knowledge. The refinement factor ranges in $[-1, 1]$ via a $tanh$ activation function.
$\epsilon$ is the control factor which prevents the model from being confused by big refinement. This factor is set to $0.1$ in our experiments.

\begin{figure}
\centering
\subfigure{
\includegraphics[width=0.6\linewidth]{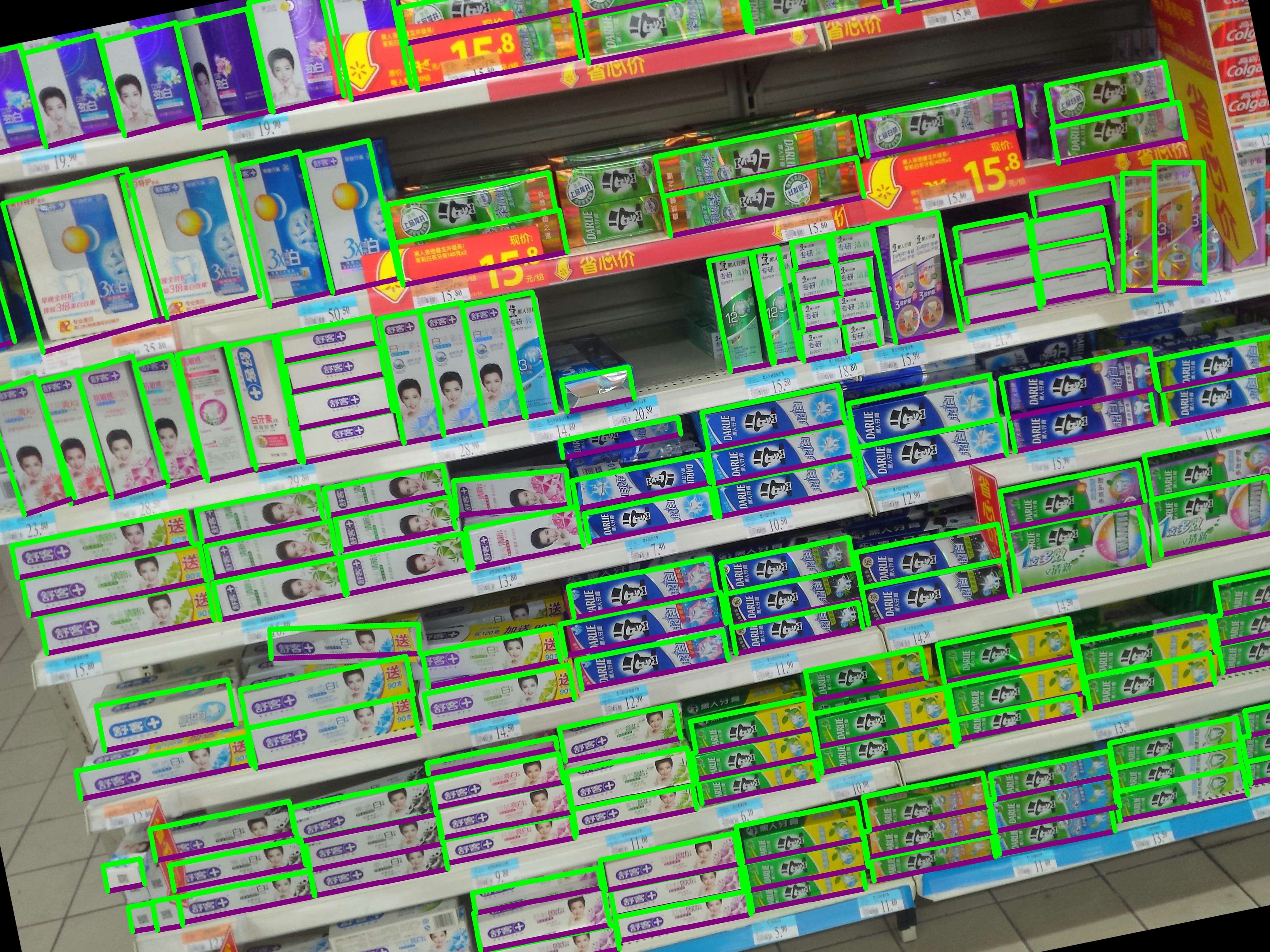}
\label{fig:rsku_0}}
\subfigure{
\includegraphics[width=0.34\linewidth]{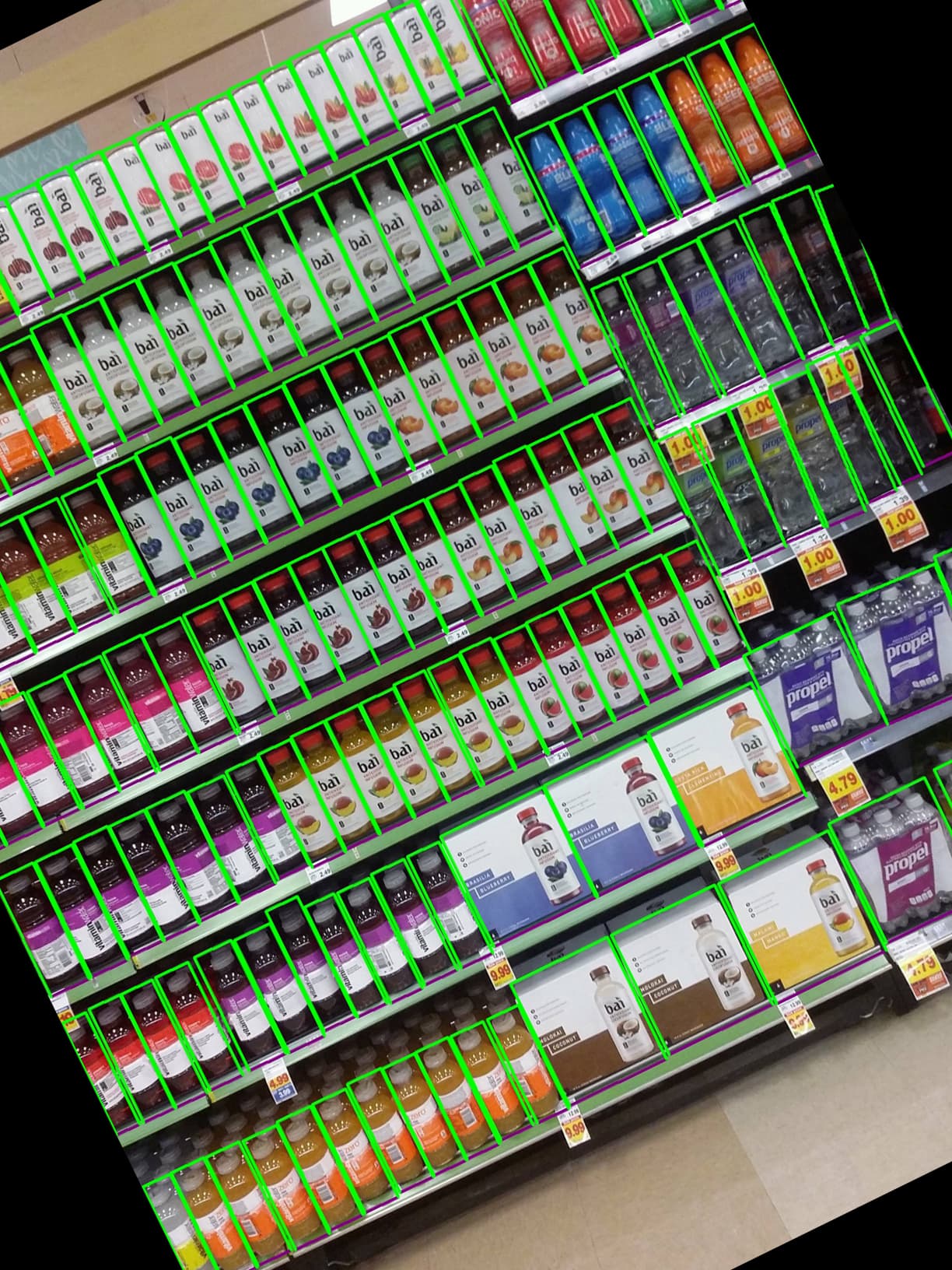}
\label{fig:rsku_1}}
\caption{Example images with annotated oriented bounding boxes in our SKU110K-R dataset.}
\label{fig:rsku110k}
\end{figure}

\subsection{SKU110K-R Dataset}
\label{sec:dataset}
Our SKU110K-R dataset is an extended version of SKU110K~\cite{goldman2019precise}.
The original SKU110K dataset contains $11,762$ images in total ($8,233$ for training, $588$ for validation, and $2,941$ for testing) and $1,733,678$ instances.
The images are collected from thousands of supermarket stores and are of various scales, viewing angles, lighting conditions, and noise levels. All the images are resized into a resolution of one megapixel.
Most of the instances in the dataset are tightly packed and typically of a certain orientation in the rage of [$-$15$^\circ$, 15$^\circ$].
To enrich the dataset, we perform data augmentation by rotating the images by six different angles, i.e., -45$^\circ$, -30$^\circ$, -15$^\circ$, 15$^\circ$, 30$^\circ$, and 45$^\circ$.
Then, we annotate the \emph{oriented bounding box} for each instance via crowdsourcing to obtain our SKU110K-R dataset.
Please refer to our supplementary materials for more details about SKU110-R.

\section{Experiments}
\label{sec:exp}
\begin{table*}
\centering
\resizebox{\textwidth}{!}{
\begin{tabular}{lccccccccccccccccc}
\toprule
Method &  PL &  BD &  BR &  GTF &  SV &  LV &  SH &  TC &  BC &  ST &  SBF &  RA &  HA &  SP &  HC &  mAP\\ \hline
\textbf{one-stage method} \\ \hline
SSD~\cite{Liu:2016:SSD} & 39.83 & 9.09 & 0.64 & 13.18 & 0.26 & 0.39 & 1.11 & 16.24 & 27.57 & 9.23 & 27.16 & 9.09 & 3.03 & 1.05 & 1.01 & 10.59 \\

YOLOv2~\cite{Redmon:2017:Yolo9000} & 39.57 & 20.29 & 36.58 & 23.42 & 8.85 & 2.09 & 4.82 & 44.34 & 38.35 & 34.65 & 16.02 & 37.62 & 47.23 & 25.5 & 7.45 & 21.39 \\

FR-O~\cite{xia2018dota} & 79.42 & 44.13 & 17.7 & 64.05 & 35.3 & 38.02 & 37.16 & 89.41 & 69.64 & 59.28 & 50.3 & 52.91 & 47.89 & 47.4 & 46.3 & 54.13 \\ \hline
\textbf{two-stage method} \\ \hline
ICN~\cite{azimi2018towards}  & 81.40 & 74.30 & 47.70 & 70.30 & 64.90 & 67.80 & 70.00 & 90.80 & 79.10 & 78.20 & 53.60 & 62.90 & 67.00 & 64.20 & 50.20 & 68.20 \\

R-DFPN~\cite{yang2018automatic} & 80.92 & 65.82 & 33.77 & 58.94 & 55.77 & 50.94 & 54.78 & 90.33 & 66.34 & 68.66 & 48.73 & 51.76 & 55.10 & 51.32 & 35.88 & 57.94 \\

R$^2$CNN~\cite{jiang2017r2cnn} & 80.94 & 65.67 & 35.34 & 67.44 & 59.92 & 50.91 & 55.81 & 90.67 & 66.92 & 72.39 & 55.06 & 52.23 & 55.14 & 53.35 & 48.22 & 60.67 \\

RRPN~\cite{ma2018arbitrary} & 88.52 & 71.20 & 31.66 & 59.30 & 51.85 & 56.19 & 57.25 & 90.81 & 72.84 & 67.38 & 56.69 & 52.84 & 53.08 & 51.94 & 53.58 & 61.01 \\

RoI-Transformer$^{*}$~\cite{ding2019learning} & 88.64 & 78.52 & 43.44 &  \textbf{75.92} & 68.81 &  73.6 & 83.59 & 90.74 & 77.27 & 81.46 & 58.39 & 53.54 & 62.83 & 58.93 & 47.67 & 69.56 \\
SCRDet~\cite{yang2019scrdet}  & 89.41 & 78.83 & 50.02 & 65.59 & 69.96 & 57.63 & 72.26 & 90.73 & 81.41 & 84.39 & 52.76 & 63.62 & 62.01 & 67.62 & 61.16 & 69.83 \\
SCRDet$^{*}$~\cite{yang2019scrdet}  & \textbf{89.9}8 & 80.65 & \textbf{52.09} & 68.36 & 68.36 &  60.32 & 72.41 & \textbf{90.85} & \textbf{87.94} & \textbf{86.86} & \textbf{65.02} & \textbf{66.68} & 66.25 & 68.24 & \textbf{65.21} & 72.61 \\
\hline
\textbf{anchor-free method} \\ \hline
baseline~\cite{zhou2019objects} & 89.02 & 69.71 & 37.62 & 63.42 & 65.23 & 63.74 & 77.28 & 90.51 & 79.24 & 77.93 & 44.83 & 54.64 & 55.93 & 61.11 & 45.71 & 65.04\\
baseline$^{**}$~\cite{zhou2019objects} & 89.56 & 79.83 & 43.8 & 66.54 & 65.58 & 66.09 & 83.11 & 90.72 & 83.72 & 84.3 & 55.62 & 58.71 & 62.48 & 68.33 & 50.77 & 69.95\\
DRN (Ours) & 88.91 & 80.222 & 43.52 & 63.35 & 73.48 & 70.69 & 84.94 & 90.14 & 83.85 & 84.11 & 50.12 & 58.41 & 67.62 & 68.60 & 52.50 & 70.70\\
DRN$^*$ (Ours) & 89.45 & {\bf83.16} & 48.98 & 62.24 & 70.63 & 74.25 & 83.99 & 90.73 & 84.60 & 85.35 & 55.76 & 60.79 & {\bf71.56} &68.82 & 63.92 & 72.95\\
DRN$^{**}$ (Ours) & 89.71 & 82.34 & 47.22 & 64.10 & \textbf{76.22} & \textbf{74.43} & {\bf85.84} & 90.57 & 86.18 & 84.89 & 57.65 & 61.93 & 69.30 & {\bf69.63} & 58.48 & {\bf73.23}\\
\hline
\bottomrule
\end{tabular}}
\caption{Evaluation results of the OBB task on the DOTA dataset. The category names are abbreviated as follows: PL-PLane, BD-Baseball Diamond, BR-BRidge, GTF-Ground Field Track, SV-Small Vehicle, LV-Large Vehicle, SH-SHip, TC-Tennis Court, BC-Basketball Court, ST-Storage Tank, SBF-Soccer-Ball Field, RA-RoundAbout, HA-Harbor, SF-Swimming Pool, and HC-HeliCopter. $(\cdot)^{*}$ represents testing in multi-scale, and $(\cdot)^{**}$ represents testing with both flip and multi-scale. The other results of our approach are all without any test augmentation.}
\label{tab:OBB_HBB}
\end{table*}
\subsection{Experimental Setup}
\paragraph{Dataset.}
We conduct experiments on three datasets, i.e., DOTA~\cite{xia2018dota}, HRSC2016~\cite{liu2016ship}, and our own SKU110K-R (Sec.~\ref{sec:dataset}).
The DOTA dataset contains $2,806$ images and covers 15 object categories. It is mainly used for object detection in aerial images with annotations of oriented bounding boxes.
The objects are of various scales, orientations, and shapes.
Before training, we crop a series of patches of the same resolution $1024 \times 1024$ from the original images with a stride of $924$\nothing{ using the official development kit} and get about $25000$ patches.
To alleviate the class imbalance, we perform data augmentation by random rotation for those categories with very few samples, and finally obtain approximately $40000$ patches in total.
The HRSC2016 dataset contains $1061$ aerial images and more than $20$ categories of ships in various appearance, scales, and orientations.
The training, validation, and test sets include $436$, $181$, and $444$ images, respectively.
We did not conduct any data augmentation on this dataset.

\nothing{
The benchmark SKU110K is a new dataset containing images of supermarket shelves where the items are in tightly packed and efficient arrangements. \xingjia{describe our re-labeled dataset instead of the original one} The training split consists of $70\%$ of images ($8,233$ images) and associated $1,210,431$ rotated bounding boxes. $5\%$ of the images (with their $90,968$ bounding boxes) are used for validation. The rest, $2,914$ images ($432,312$ bounding boxes) are used for testing. We relabel a oriented bounding box for each instance.
}

For the DOTA and HRSC2016 datasets, we use the same mAP calculation as PASCAL VOC~\cite{everingham2010pascal}.
For SKU110K and SKU110K-R, we use the same evaluation method as COCO~\cite{lin2014microsoft}, which reported an mean average precision (mAP) at IoU $=0.5 :0.05 :0.95$.
Moreover, we report AP at IoU $=0.75$ (AP$_{75}$) and average recall $300$ (AR$_{300}$) at IoU $= 0.5 : 0.05 : 0.95$ ($300$ is the maximal number of objects) following Goldman \etal~\cite{goldman2019precise}.

\paragraph{Implementation details.}
We use an hourglass-104 network as the backbone.
To implement RCL, we use the released code of DCNV2~\cite{zhu2019deformable} and replace the original predicted offset with the offset deduced from the predicted angle in Eqn.~\ref{Equ:off}.

The input resolutions of DOTA, HRSC2016, and SKU110K-R are $1024\times 1024$, $768 \times 768$, and $768 \times 768$, respectively.
We used random scaling (in the range of $[0.7, \, 1.3]$), random flipping, and color jittering for data augmentation.
For DOTA and HRSC, the models are trained with $140$ epochs in total.
The learning rate is reduced by a factor of $10$ after the $90$th and the $120$th epochs from an initial value of $4e-4$ to $4e-6$ finally.
For SKU110K-R, we train for $25$ epochs, with a learning rate of $2e-4$ which is decreased by 10 at the $20$ epoch.
We use Adam~\cite{kingma2014adam} as the optimizer and set the batch size to $8$.
For improved convergence, we calculate the offsets from target angles instead of predicted ones during the training phase.
\begin{table*}
\centering
\begin{tabular}{lccccccc|c}
\toprule
    Method &  CP~\cite{liu2017rotated} &  BL$_2$~\cite{liu2017rotated} &  RC$_1$~\cite{liu2017rotated} &  RC$_2$~\cite{liu2017rotated} &  R$^2$PN~\cite{zhang2018toward} &  RRD~\cite{liao2018rotation} &RoI Trans~\cite{ding2019learning} &  Ours  \\ \hline\hline
    mAP & 55.7 & 69.6 & 75.7 & 75.7 & 79.6 & 84.3 & 86.2 & $\textbf{92.7}$ \\
\bottomrule
\end{tabular}
\caption{Evaluation results on the HRSC2016 dataset.}
\label{tab:hrsc}
\end{table*}

We deduce the offset in RCL using predicted angles at the test time.
As set in CenterNet, we adopt three levels of test augmentation.
First, we evaluate our method without any augmentation.
Then, we add multi-scale testing with $(0.5,1.0,1.5)$. To merge the detection, we adopt a variant of Soft-NMS~\cite{bodla2017soft} that faces oriented bounding boxes (angle-softnms).
Specifically, we use the linear method to adjust the score value, set the IoU threshold, and suppress the threshold to $0.5$ and $0.03$, respectively.
Lastly, we add horizontal flipping and average the network predictions before decoding oriented bounding boxes.

\subsection{Experimental Results}
Table~\ref{tab:OBB_HBB} shows quantitative results comparing our approach with state-of-the-art methods on the DOTA test set for the oriented bounding box (OBB) task.
Other methods are all anchor-based and most of them are based on the framework of Faster R-CNN~\cite{ren2015faster}.
By contrast, we follow an anchor-free paradigm and demonstrate comparable results with SCRDet~\cite{yang2019scrdet}\nothing{, demonstrating the competitive edge of our proposed method}.
Compared to the baseline, our method ahchieves a remarkable gain of $3.3\%$ in terms of mAP.

\begin{table}
\centering
\begin{tabular}{cccc}
\toprule
 Method&mAP &AP$_{50}$ &  AP$_{75}$   \\ \hline\hline
 Baseline & 63.5 & \textbf{92.3}  & 75.4 \\
 Ours &\textbf{65.6} & 92.0 &\textbf{77.8} \\
\bottomrule
\end{tabular}
\caption{Comparison of our method with the baseline on the HRSC2016 dataset in COCO fashion.
}
\label{tab:hrsc_coco}
\end{table}

Table~\ref{tab:hrsc} shows the results on HRSC2016 in Pascal VOC fashion.
Our method achieves a significant gain of $6.4\%$ in terms of mAP.
Such improvement indicates that the proposed FSM effectively addresses the misalignment issue by adjusting the receptive fields adaptively.
We further show evaluation results on COCO fashion in Table~\ref{tab:hrsc_coco}\nothing{ to compare our method with the baseline}.
Our method provides $1.9\%$ mAP gain.
Moreover, as the IoU increases, our method improves.
Fig.~\ref{fig:vis_results} shows some qualitative results on DOTA and HRSC2016 datasets using our method.

\begin{table}
\centering
\resizebox{1.0\linewidth}{!}{
    \begin{tabular}{llccc}
    	\toprule
    	Dataset & Method & mAP & AP$_{75}$ & AR$_{300}$\\ \hline
        \multirow{7}{*}{SKU110K}
    	& Faster-RCNN~\cite{ren2015faster} & 4.5 & 1.0 & 6.6 \\
    	& YOLO9000~\cite{Redmon:2017:Yolo9000}  & 9.4 & 7.3 &  11.1 \\
    	& RetinaNet~\cite{lin2017focal} & 45.5 & 38.9 & 53.0 \\
    	& RetinaNet with EM-Merger~\cite{goldman2019precise}  &  49.2 & 55.6 &55.4 \\
    	& YoloV3~\cite{redmon2018yolov3} &  55.4& 76.8 & 56.2 \\
    	& Baseline &  55.8& 62.8 & 62.5 \\
    	& Ours & {\bf56.9}& \textbf{64.0} & \textbf{63.5} \\
    \hline
    	\multirow{4}{*}{SKU110K-R}
    	& YoloV3-Rotate & 49.1 & 51.1 & 58.2 \\
    	& CenterNet-4point$^\dagger$~\cite{zhou2019objects} & 34.3 & 19.6 & 42.2 \\
    	& CenterNet$^\dagger$~\cite{zhou2019objects} & 54.7 & 61.1 & 62.2 \\
    	& Baseline & 54.4 & 60.6 & 61.6 \\
    	& Ours & \bf55.9 & \textbf{63.1} & \textbf{63.3} \\
    	\bottomrule
    \end{tabular}
}
\caption{Evaluation results on SKU110K and SKU110K-R.}
\label{tab:sku110k}
\end{table}

Table~\ref{tab:sku110k} shows the results on SKU110K-R and SKU110K.
For oriented object detection, we reimplement YoloV3~\cite{redmon2018yolov3} by introducing angle prediction.
CenterNet-4point$^\dagger$ represents regressing the four corners of each bounding box, and CenterNet$^\dagger$ indicates that we add center pooling and DCN~\cite{dai2017deformable} to our baseline.
We improve the mAP by $1.5\%$ and also report superior results on the original SKU110K dataset.
These numbers further demonstrate the effectiveness of our proposed DRN.
\begin{table}
\centering
\begin{tabular}{lccccc}
	\toprule
      Method &MK &DCN &ROT & AP$_{50}$ & AP$_{75}$  \\ \hline
      Baseline & & & & 63.4 & 34.6 \\ \hline
     \multirow{9}{*}{FSM}
      &$33$ & & & 63.3 & 34.5 \\ 
      & $33$ &$\checkmark$ & & 63.5 & 34.8 \\
      & $33$ & &$\checkmark$ & 63.9 & 35.1 \\
      & $33, 13$ & & & 63.5 & 34.7\\
      & $33,13$ &$\checkmark$ & & 63.6 &34.9 \\
      & $33, 13$ & &$\checkmark$ & 64.2 &35.4 \\
      & $33,13,31$ & & & 63.7 & 34.8\\
      & $33,13,31$ &$\checkmark$ & & 63.9 & 35.2 \\
      & $33,13,31$ & &$\checkmark$ &\textbf{64.4} & \textbf{35.7} \\
	\bottomrule
\end{tabular}
\caption{Ablation studies about FSM on the DOTA validation set. MK denotes the multiple kernels used in FSM. $33$, $13$, and $31$ represent kernel sizes of $(3,3), (1,3)$ and $(3,1)$, respectively. DCN and ROT are the deformable and rotation convolution layers.}
\label{tab:fsm}
\end{table}

\subsection{Ablation Study}
We conduct a series of ablation studies on the DOTA validation set and report quantitative results in COCO fashion 
to verify the effectiveness of our method.
We use the hourglass-52 as our backbone in this section.
\newcommand\exampleresultwidth{0.24}
\begin{figure*}
	\centering
		\includegraphics[width=\exampleresultwidth\linewidth]{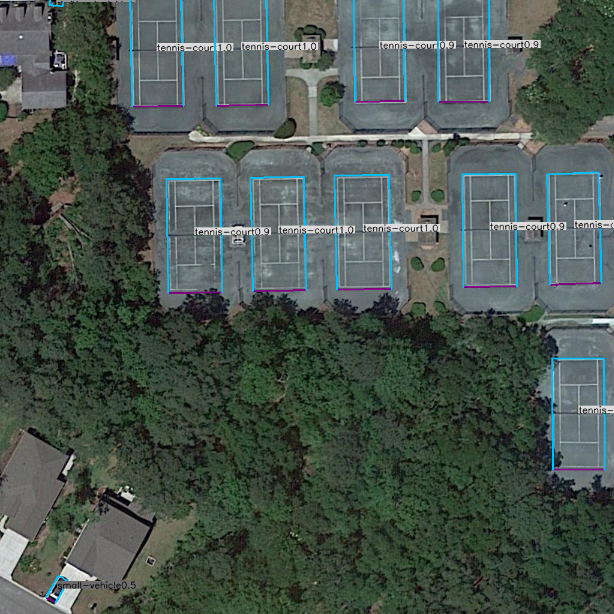}
		\includegraphics[width=\exampleresultwidth\linewidth]{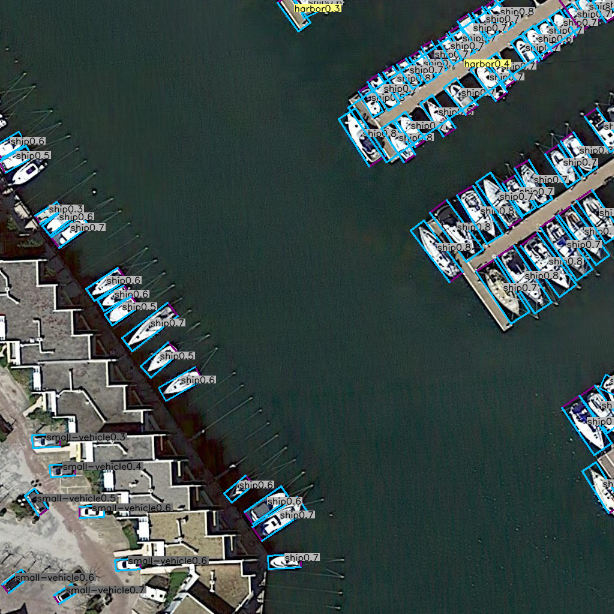}
		\includegraphics[width=\exampleresultwidth\linewidth]{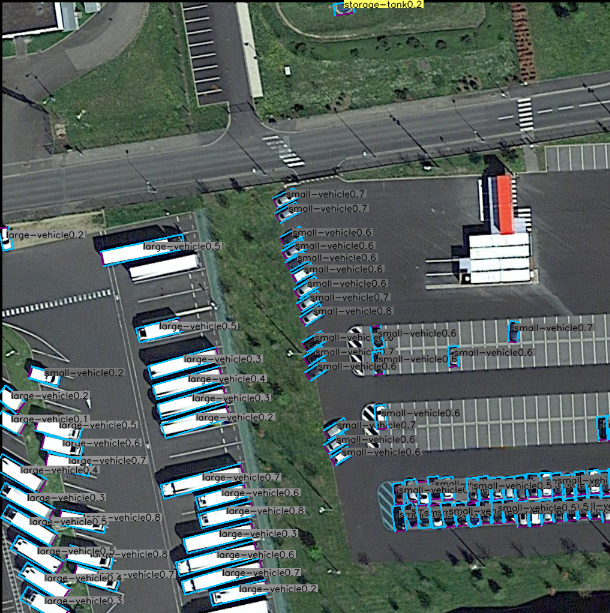}
		\includegraphics[width=\exampleresultwidth\linewidth]{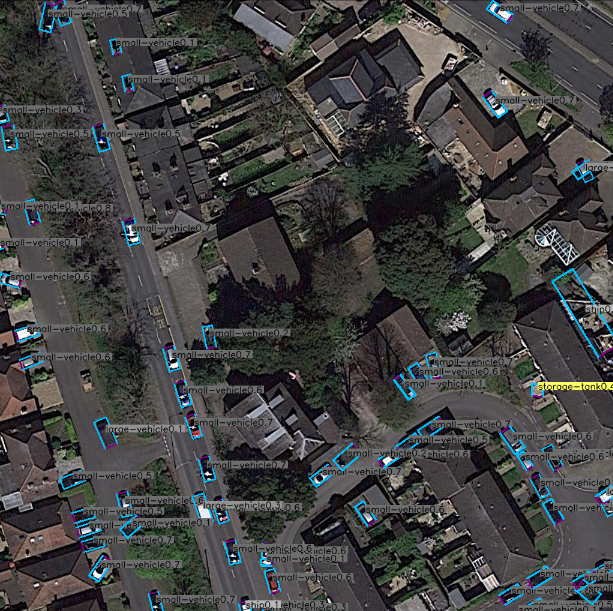}

		\vspace{3pt}
		\includegraphics[width=\exampleresultwidth\linewidth]{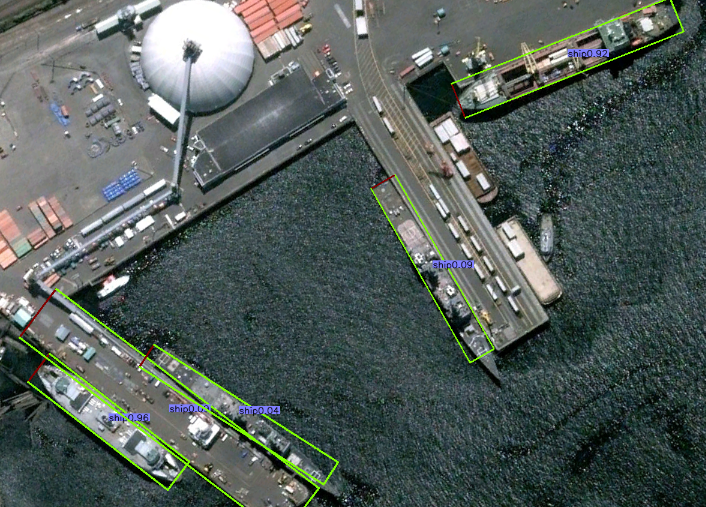}
		\includegraphics[width=\exampleresultwidth\linewidth]{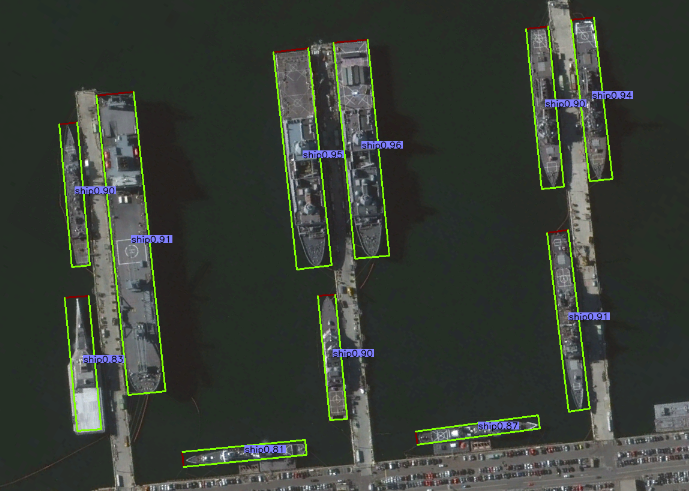}
		\includegraphics[width=\exampleresultwidth\linewidth]{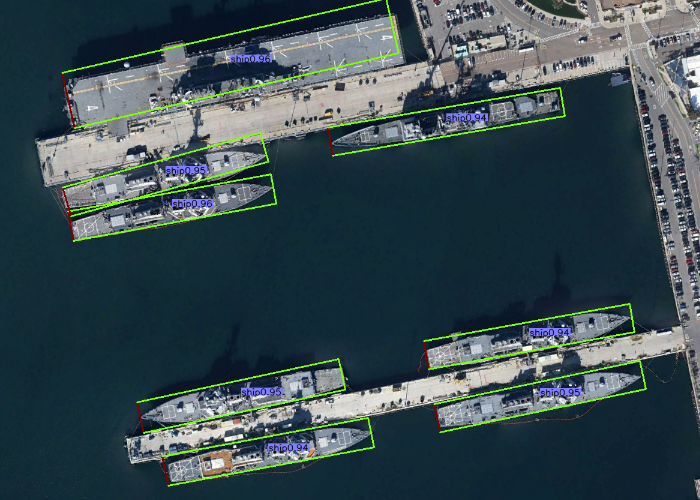}
		\includegraphics[width=\exampleresultwidth\linewidth]{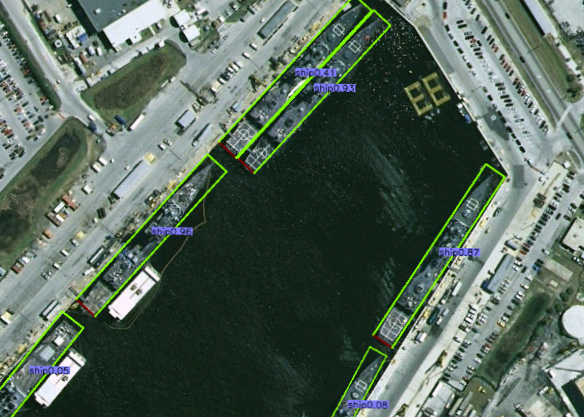}
	\caption{Example detection results of our method. The \textbf{top} row is from DOTA while and the \textbf{bottom} row is from HRSC2016.}
	\vspace{-1mm}
	\label{fig:vis_results}
\end{figure*}

Our FSM aims to select compact receptive fields for each object adaptively.
To match the objects as much as possible, we set up three shapes of kernels, i.e., square, flat, and slender rectangles.
Table~\ref{tab:fsm} shows the results when we use different settings.
The first row is the baseline.
We first construct the FSM with only one branch by using a $3\times 3$ kernel and shield the RCL.
This setting achieves almost the same results as the baseline since our network is the same as the baseline, except for the addition of one convolution layer before the head branches.
When we add the RCL, some improvement ($0.5\%$) is observed because the RCL enables the neurons to adjust receptive fields by rotation.
Next, we add a flat kernel ($1\times 3$) and the model demonstrates improved performance.
Lastly, we add a slender kernel ($3\times 1$) and the model shows consistent gains.
The FSM with three splits enables the neurons to adjust receptive fields in two degrees of freedom, namely, shape and rotation.
A slight improvement of a few more flat-shaped objects is observed when $1\times 3$ kernel is added.
To further reveal the effectiveness of FSM, we visualize the attention map in FSM.
Details are available in our supplementary materials.
In our experiments, we set up simple kernels to demonstrate the effectiveness of FSM and leave the design of more complex kernel shapes as future work.

\begin{table}
\centering
{
\begin{tabular}{lcccc}
	\toprule
     Method &Acc &Rec & AP$_{50}$ &AP$_{75}$ \\ \hline
	 Baseline &0.21 &0.89 & 63.4 & 34.6 \\
     DRH-C & \textbf{0.27}  & \textbf{0.95} & \textbf{64.1} & \textbf{35.2}\\
	\bottomrule
\end{tabular}
}
\caption{Evaluation results on the validation partition of the DOTA dataset using DRH-C.}
\label{tab:drmc}
\end{table}

\begin{table}
\centering
\begin{tabular}{lccccc}
	\toprule
     \multirow{2}{*}{Method} &\multicolumn{3}{c}{$L_1$} &\multirow{2}{*}{AP$_{50}$} &\multirow{2}{*}{AP$_{75}$} \\  
     &scale & angle & offset & \\ \hline
	 Baseline & 5.34 & 0.21 & 0.39 &63.4 &34.6 \\ \hline
	 \multirow{4}{*}{DRH-R}
       &4.12 & - &- &64.1 &35.2 \\ 
       & - & 0.19 &- &63.5 &34.8 \\ 
       & - & - &0.36 &63.4 &34.5 \\ 
       & \textbf{4.10} & \textbf{0.18} & \textbf{0.35} &\textbf{64.1} &\textbf{35.3} \\
	\bottomrule
\end{tabular}
\caption{Evaluation results on the DOTA validation set using DRH-R.}
\label{tab:drmr}
\end{table}

\begin{table}
\centering
\begin{tabular}{lccccc}
    \toprule
      Method  &$T_{test}$ & $Params$ & AP$_{50}$ & AP$_{75}$  \\ \hline
      Baseline &0.078s & - & 63.4 & 34.6 \\ \hline
      +FSM &0.086s &+~0.1M & 64.4 & 35.7 \\ 
      +DRH-C &0.095s &+~0.03M  & 65.0 & 36.3\\
      +DRH-R &0.102s &+~0.03M & 65.7 &36.9 \\
     \bottomrule
\end{tabular}
\caption{Comparison of our method with the baseline in terms of speed, complexity, and accuracy. The timing information is measured using images resolution $1024 \times 1024$ on a single NVIDIA Tesla V100. The time of post-processing (i.e., NMS) is not included.}
\label{tab:sz_sp}
\end{table}

To model the uniqueness and particularity of each object and empower the network to handle flexible samples, we design two DRHs for classification and regression tasks.
For the classifier, we report the accuracy (Acc), recall (Rec), and AP to reveal the quality of center point prediction.
Specifically, we select the top $300$ points as the predicted object centers for each image in our experiments.

Table~\ref{tab:drmc} shows the results of the ablation study on DRH-C.
The performance of the classifier is considerably improved when DRH-C is introduced.
Specifically, the Acc and Rec are increased from $0.21$ to $0.32$ and from $0.81$ to $0.89$, respectively.
For the detection, the DHR-C provides $0.7\%$ AP$_{50}$ and $0.6\%$ AP$_{75}$ gains.
In Table~\ref{tab:drmr}, to evaluate the impact of DRH-R, we report the prediction errors, AP$_{50}$, and AP$_{75}$ when we replace the original heads with our DRH-R for scale, angle, and offset regression.
We use the standard $L_1$ distance between the predicted and ground-truth values to measure the errors.
The first three rows in DRH-R show results when we replace the corresponding single head with DRH-R.
Our DHR-R provides consistent improvement albeit slight on angle and offset regression tasks. The reason is that these two tasks are relatively easy and have almost achieved the optimal point.
On scale regression tasks, DRH-R reduces $L_1$ error by $1.24$ and improves AP$_{50}$ and AP$_{75}$ by  $0.7\%$ and $0.6\%$, respectively.
Table~\ref{tab:sz_sp} compares our method with the baseline in terms of the average time to process an image, numbers of model parameters, as well as model performance (AP$_{50}$ and AP$_{75}$).
Our method has achieved remarkable improvement over the baseline with very limited increased number of parameters.
Here, we only apply DRH-R on the scale head.

\section{Conclusion}
\label{sec:conc}
In this work, we present a unified framework for oriented and densely packed object detection.
To adjust receptive fields of neurons in accordance with object shapes and orientations, we propose an FSM to aggregate information and thus address the misalignment issue between receptive fields and various objects.
We further present DRH-C and DRH-R to refine the prediction dynamically, thereby alleviating the contradiction between the model equipped by generic knowledge and specific objects.
In addition, we relabel SKU110K with oriented bounding boxes and obtain a new dataset, called SKU110K-R to facilitate the development of detection models on oriented and densely packed objects.
We conduct extensive experiments to show that our method achieves consistent gains across multiple datasets in comparison with baseline approaches.
In the future, we plan to explore a more effective mechanism of dynamic models and investigate oriented object detection in few-shot settings.

\small{
\paragraph{Acknowledgment.}
This work was supported by National Key R\&D Program of China under no. 2018YFC0807500, and by National Natural Science Foundation of China under nos. 61832016, 61672520 and 61720106006, and by CASIA-Tencent Youtu joint research project.
}

{\small
\bibliographystyle{ieee_fullname}
\bibliography{drn}

\begin{thebibliography}{10}\itemsep=-1pt

\bibitem{azimi2018towards}
Seyed~Majid Azimi, Eleonora Vig, Reza Bahmanyar, Marco K{\"o}rner, and Peter
  Reinartz.
\newblock Towards multi-class object detection in unconstrained remote sensing
  imagery.
\newblock In {\em Asian Conference on Computer Vision}, pages 150--165.
  Springer, 2018.

\bibitem{bodla2017soft}
Navaneeth Bodla, Bharat Singh, Rama Chellappa, and Larry~S Davis.
\newblock {Soft-NMS -- Improving Object Detection With One Line of Code}.
\newblock In {\em Proceedings of the IEEE International Conference on Computer
  Vision}, pages 5561--5569, 2017.

\bibitem{Borji:2019:SOD}
Ali Borji, Ming-Ming Cheng, Qibin Hou, Huaizu Jiang, and Jia Li.
\newblock Salient object detection: A survey.
\newblock {\em Computational Visual Media}, 5(2):117--150, 2019.

\bibitem{dai2017deformable}
Jifeng Dai, Haozhi Qi, Yuwen Xiong, Yi Li, Guodong Zhang, Han Hu, and Yichen
  Wei.
\newblock Deformable convolutional networks.
\newblock In {\em Proceedings of the IEEE International Conference on Computer
  Vision}, pages 764--773, 2017.

\bibitem{ding2019learning}
Jian Ding, Nan Xue, Yang Long, Gui-Song Xia, and Qikai Lu.
\newblock Learning roi transformer for oriented object detection in aerial
  images.
\newblock In {\em Proceedings of the IEEE Conference on Computer Vision and
  Pattern Recognition}, pages 2849--2858, 2019.

\bibitem{everingham2010pascal}
Mark Everingham, Luc Van~Gool, Christopher~KI Williams, John Winn, and Andrew
  Zisserman.
\newblock {The PASCAL Visual Object Classes (VOC) Challenge}.
\newblock {\em International Journal of Computer Vision}, 88(2):303--338, 2010.

\bibitem{girshick2015fast}
Ross Girshick.
\newblock {Fast R-CNN}.
\newblock In {\em Proceedings of the IEEE International Conference on Computer
  Vision}, pages 1440--1448, 2015.

\bibitem{girshick2014rich}
Ross Girshick, Jeff Donahue, Trevor Darrell, and Jitendra Malik.
\newblock Rich feature hierarchies for accurate object detection and semantic
  segmentation.
\newblock In {\em Proceedings of the IEEE Conference on Computer Vision and
  Pattern Recognition}, pages 580--587, 2014.

\bibitem{goldman2019precise}
Eran Goldman, Roei Herzig, Aviv Eisenschtat, Jacob Goldberger, and Tal Hassner.
\newblock Precise detection in densely packed scenes.
\newblock In {\em Proceedings of the IEEE Conference on Computer Vision and
  Pattern Recognition}, pages 5227--5236, 2019.

\bibitem{hsieh2017drone}
Meng-Ru Hsieh, Yen-Liang Lin, and Winston~H Hsu.
\newblock Drone-based object counting by spatially regularized regional
  proposal network.
\newblock In {\em Proceedings of the IEEE International Conference on Computer
  Vision}, pages 4145--4153, 2017.

\bibitem{hu2018squeeze}
Jie Hu, Li Shen, and Gang Sun.
\newblock Squeeze-and-excitation networks.
\newblock In {\em Proceedings of the IEEE Conference on Computer Vision and
  Pattern Recognition}, pages 7132--7141, 2018.

\bibitem{ioffe2015batch}
Sergey Ioffe and Christian Szegedy.
\newblock Batch normalization: Accelerating deep network training by reducing
  internal covariate shift.
\newblock {\em arXiv preprint arXiv:1502.03167}, 2015.

\bibitem{jaderberg2015spatial}
Max Jaderberg, Karen Simonyan, Andrew Zisserman, and Koray Kavukcuoglu.
\newblock Spatial transformer networks.
\newblock In {\em Advances in Neural Information Processing Systems}, pages
  2017--2025, 2015.

\bibitem{jeon2017active}
Yunho Jeon and Junmo Kim.
\newblock Active convolution: Learning the shape of convolution for image
  classification.
\newblock In {\em Proceedings of the IEEE Conference on Computer Vision and
  Pattern Recognition}, pages 4201--4209, 2017.

\bibitem{jia2016dynamic}
Xu Jia, Bert De~Brabandere, Tinne Tuytelaars, and Luc~V Gool.
\newblock Dynamic filter networks.
\newblock In {\em Advances in Neural Information Processing Systems}, pages
  667--675, 2016.

\bibitem{jiang2017r2cnn}
Yingying Jiang, Xiangyu Zhu, Xiaobing Wang, Shuli Yang, Wei Li, Hua Wang, Pei
  Fu, and Zhenbo Luo.
\newblock {R2CNN: Rotational Region CNN for Orientation Robust Scene Text
  Detection}.
\newblock {\em arXiv preprint arXiv:1706.09579}, 2017.

\bibitem{kingma2014adam}
Diederik~P Kingma and Jimmy Ba.
\newblock Adam: A method for stochastic optimization.
\newblock {\em arXiv preprint arXiv:1412.6980}, 2014.

\bibitem{law2018cornernet}
Hei Law and Jia Deng.
\newblock Cornernet: Detecting objects as paired keypoints.
\newblock In {\em Proceedings of the European Conference on Computer Vision},
  pages 734--750, 2018.

\bibitem{li2019dynamic}
Shuai Li, Lingxiao Yang, Jianqiang Huang, Xian-Sheng Hua, and Lei Zhang.
\newblock Dynamic anchor feature selection for single-shot object detection.
\newblock In {\em Proceedings of the IEEE International Conference on Computer
  Vision}, pages 6609--6618, 2019.

\bibitem{li2019selective}
Xiang Li, Wenhai Wang, Xiaolin Hu, and Jian Yang.
\newblock Selective kernel networks.
\newblock In {\em Proceedings of the IEEE Conference on Computer Vision and
  Pattern Recognition}, pages 510--519, 2019.

\bibitem{liao2018rotation}
Minghui Liao, Zhen Zhu, Baoguang Shi, Gui-song Xia, and Xiang Bai.
\newblock Rotation-sensitive regression for oriented scene text detection.
\newblock In {\em Proceedings of the IEEE Conference on Computer Vision and
  Pattern Recognition}, pages 5909--5918, 2018.

\bibitem{lin2017feature}
Tsung-Yi Lin, Piotr Doll{\'a}r, Ross Girshick, Kaiming He, Bharath Hariharan,
  and Serge Belongie.
\newblock Feature pyramid networks for object detection.
\newblock In {\em Proceedings of the IEEE Conference on Computer Vision and
  Pattern Recognition}, pages 2117--2125, 2017.

\bibitem{lin2017focal}
Tsung-Yi Lin, Priya Goyal, Ross Girshick, Kaiming He, and Piotr Doll{\'a}r.
\newblock Focal loss for dense object detection.
\newblock In {\em Proceedings of the IEEE International Conference on Computer
  Vision}, pages 2980--2988, 2017.

\bibitem{Lin:2014:COCO}
Tsung-Yi Lin, Michael Maire, Serge Belongie, James Hays, Pietro Perona, Deva
  Ramanan, Piotr Doll{\'a}r, and C.~Lawrence Zitnick.
\newblock Microsoft {COCO}: Common objects in context.
\newblock In {\em Proceedings of the European Conference on Computer Vision},
  pages 740--755, Cham, 2014.

\bibitem{lin2014microsoft}
Tsung-Yi Lin, Michael Maire, Serge Belongie, James Hays, Pietro Perona, Deva
  Ramanan, Piotr Doll{\'a}r, and C~Lawrence Zitnick.
\newblock Microsoft coco: Common objects in context.
\newblock In {\em European Conference on Computer Vision}, pages 740--755,
  2014.

\bibitem{liu2015fast}
Kang Liu and Gellert Mattyus.
\newblock Fast multiclass vehicle detection on aerial images.
\newblock {\em IEEE Geoscience and Remote Sensing Letters}, 12(9):1938--1942,
  2015.

\bibitem{Liu:2016:SSD}
Wei Liu, Dragomir Anguelov, Dumitru Erhan, Christian Szegedy, Scott Reed,
  Cheng-Yang Fu, and Alexander~C Berg.
\newblock {SSD}: Single shot multibox detector.
\newblock In {\em European Conference on Computer Vision}, pages 21--37, 2016.

\bibitem{liu2017rotated}
Zikun Liu, Jingao Hu, Lubin Weng, and Yiping Yang.
\newblock Rotated region based cnn for ship detection.
\newblock In {\em 2017 IEEE International Conference on Image Processing},
  pages 900--904, 2017.

\bibitem{liu2016ship}
Zikun Liu, Hongzhen Wang, Lubin Weng, and Yiping Yang.
\newblock Ship rotated bounding box space for ship extraction from
  high-resolution optical satellite images with complex backgrounds.
\newblock {\em IEEE Geoscience and Remote Sensing Letters}, 13(8):1074--1078,
  2016.

\bibitem{ma2018arbitrary}
Jianqi Ma, Weiyuan Shao, Hao Ye, Li Wang, Hong Wang, Yingbin Zheng, and
  Xiangyang Xue.
\newblock Arbitrary-oriented scene text detection via rotation proposals.
\newblock {\em IEEE Transactions on Multimedia}, 20(11):3111--3122, 2018.

\bibitem{nair2010rectified}
Vinod Nair and Geoffrey~E Hinton.
\newblock Rectified linear units improve restricted boltzmann machines.
\newblock In {\em Proceedings of the 27th International Conference on Machine
  Learning}, pages 807--814, 2010.

\bibitem{Redmon:2016:YOLO}
Joseph Redmon, Santosh Divvala, Ross Girshick, and Ali Farhadi.
\newblock You only look once: Unified, real-time object detection.
\newblock In {\em IEEE Conference on Computer Vision and Pattern Recognition},
  pages 779--788, 2016.

\bibitem{Redmon:2017:Yolo9000}
Joseph Redmon and Ali Farhadi.
\newblock Yolo9000: Better, faster, stronger.
\newblock In {\em IEEE Conference on Computer Vision and Pattern Recognition},
  pages 6517--6525, July 2017.

\bibitem{redmon2018yolov3}
Joseph Redmon and Ali Farhadi.
\newblock Yolov3: An incremental improvement.
\newblock {\em arXiv preprint arXiv:1804.02767}, 2018.

\bibitem{ren2015faster}
Shaoqing Ren, Kaiming He, Ross Girshick, and Jian Sun.
\newblock {Faster R-CNN: Towards real-time object detection with region
  proposal networks}.
\newblock In {\em Advances in Neural Information Processing Systems}, pages
  91--99, 2015.

\bibitem{Shen:2017:DSOD}
Zhiqiang Shen, Zhuang Liu, Jianguo Li, Yu-Gang Jiang, Yurong Chen, and
  Xiangyang Xue.
\newblock Dsod: Learning deeply supervised object detectors from scratch.
\newblock In {\em Proceedings of the IEEE International Conference on Computer
  Vision}, pages 1919--1927, 2017.

\bibitem{Song:2019:TSR}
Yizhi Song, Ruochen Fan, Sharon Huang, Zhe Zhu, and Ruofeng Tong.
\newblock A three-stage real-time detector for traffic signs in large
  panoramas.
\newblock {\em Computational Visual Media}, 5(4):403--416, 2019.

\bibitem{wang2019carafe}
Jiaqi Wang, Kai Chen, Rui Xu, Ziwei Liu, Chen~Change Loy, and Dahua Lin.
\newblock Carafe: Content-aware reassembly of features.
\newblock In {\em Proceedings of the IEEE International Conference on Computer
  Vision}, pages 3007--3016, 2019.

\bibitem{woo2018cbam}
Sanghyun Woo, Jongchan Park, Joon-Young Lee, and In So~Kweon.
\newblock Cbam: Convolutional block attention module.
\newblock In {\em Proceedings of the European Conference on Computer Vision},
  pages 3--19, 2018.

\bibitem{xia2018dota}
Gui-Song Xia, Xiang Bai, Jian Ding, Zhen Zhu, Serge Belongie, Jiebo Luo, Mihai
  Datcu, Marcello Pelillo, and Liangpei Zhang.
\newblock Dota: A large-scale dataset for object detection in aerial images.
\newblock In {\em Proceedings of the IEEE Conference on Computer Vision and
  Pattern Recognition}, pages 3974--3983, 2018.

\bibitem{yang2018automatic}
Xue Yang, Hao Sun, Kun Fu, Jirui Yang, Xian Sun, Menglong Yan, and Zhi Guo.
\newblock Automatic ship detection in remote sensing images from google earth
  of complex scenes based on multiscale rotation dense feature pyramid
  networks.
\newblock {\em Remote Sensing}, 10(1):132, 2018.

\bibitem{yang2019scrdet}
Xue Yang, Jirui Yang, Junchi Yan, Yue Zhang, Tengfei Zhang, Zhi Guo, Xian Sun,
  and Kun Fu.
\newblock {SCRDet: Towards More Robust Detection for Small, Cluttered and
  Rotated Objects}.
\newblock In {\em Proceedings of the IEEE International Conference on Computer
  Vision}, pages 8232--8241, 2019.

\bibitem{zhang2018toward}
Zenghui Zhang, Weiwei Guo, Shengnan Zhu, and Wenxian Yu.
\newblock Toward arbitrary-oriented ship detection with rotated region proposal
  and discrimination networks.
\newblock {\em IEEE Geoscience and Remote Sensing Letters}, 15(11):1745--1749,
  2018.

\bibitem{zhou2019objects}
Xingyi Zhou, Dequan Wang, and Philipp Kr{\"a}henb{\"u}hl.
\newblock Objects as points.
\newblock {\em arXiv preprint arXiv:1904.07850}, 2019.

\bibitem{zhou2017oriented}
Yanzhao Zhou, Qixiang Ye, Qiang Qiu, and Jianbin Jiao.
\newblock Oriented response networks.
\newblock In {\em Proceedings of the IEEE Conference on Computer Vision and
  Pattern Recognition}, pages 519--528, 2017.

\bibitem{zhu2019feature}
Chenchen Zhu, Yihui He, and Marios Savvides.
\newblock Feature selective anchor-free module for single-shot object
  detection.
\newblock In {\em Proceedings of the IEEE Conference on Computer Vision and
  Pattern Recognition}, pages 840--849, 2019.

\bibitem{zhu2019deformable}
Xizhou Zhu, Han Hu, Stephen Lin, and Jifeng Dai.
\newblock Deformable convnets v2: More deformable, better results.
\newblock In {\em Proceedings of the IEEE Conference on Computer Vision and
  Pattern Recognition}, pages 9308--9316, 2019.

\end{thebibliography}
}

\end{document}